# A New Scope and Domain Measure Comparison Method for Global Convergence Analysis in Evolutionary Computation

Liu-Yue Luo, *Student Member, IEEE*, Zhi-Hui Zhan, *Fellow, IEEE*, Kay Chen Tan, *Fellow, IEEE*, and Jun Zhang, *Fellow*, IEEE

*Abstract*—Convergence analysis is a fundamental research topic in evolutionary computation (EC). The commonly used analysis method models the EC algorithm as a homogeneous Markov chain for analysis, which is not always suitable for different EC variants, and also sometimes causes misuse and confusion due to their complex process. In this article, we categorize the existing researches on convergence analysis in EC algorithms into stable convergence and global convergence, and then prove that the conditions for these two convergence properties are somehow mutually exclusive. Inspired by this proof, we propose a new scope and domain measure comparison (SDMC) method for analyzing the global convergence of EC algorithms and provide a rigorous proof of its necessity and sufficiency as an alternative condition. Unlike traditional methods, the SDMC method is straightforward, bypasses Markov chain modeling, and minimizes errors from misapplication as it only focuses on the measure of the algorithm's search scope. We apply SDMC to two algorithm types that are unsuitable for traditional methods, confirming its effectiveness in global convergence analysis. Furthermore, we apply the SDMC method to explore the gene targeting mechanism's impact on the global convergence in large-scale global optimization, deriving insights into how to design EC algorithms that guarantee global convergence and exploring how theoretical analysis can guide EC algorithm design.

## I. INTRODUCTION

Evolutionary computation (EC) algorithms are a class of stochastic search algorithms inspired by natural evolution. They are of great significance due to their strong optimization capabilities, wide applicability, global search properties, ease of parallelization, flexibility, and scalability [1]-[3]. EC algorithms have been widely applied in various real-world scenarios, including engineering optimization[4], machine learning [5][6], bioinformatics [7][8], transportation and logistics optimization [9], financial modeling and portfolio optimization [10][11], energy and environmental management [12][13], and healthcare [14].

Although the research on the design and application of EC algorithms has been greatly developed, there is not that much research on the theoretical analysis of EC. Even in these few theoretical analysis studies, most of them, e.g., runtime analysis [15]-[21], focus on the practical performance analysis of algorithms in specific problems. This kind of research aims to provide a more precise and detailed analysis of algorithm's behavior, revealing performance characteristics and limitations in practical applications. However, it inevitably needs to know the characteristics of the problem to be solved, whereas only limited or even no information can be known in many cases, such as black-box optimization problems. In addition, due to the complexity of the calculation, the analysis in these methods is limited to some toy model problems that are far from reality. Apart from that, there is even less theoretical analysis research on algorithms' properties regardless of the characteristics of problems.

In theoretical analysis research on EC algorithms themselves, convergence is one of the most frequently mentioned properties reflecting the optimization capabilities of the algorithms. However, the concept of convergence is not clarified across different studies in the literature. For example, when both are analyzing whether the particle swarm optimization (PSO) algorithm can converge to the global optimum with probability 1, studies in [21] and [21] reach completely different conclusions. The research in [21] claims to have proven that PSO guarantees global convergence, while the research in [23] arrives at the opposite conclusion. This is due to that the research in [21] mistakenly interpreted the convergence of the sequence composed of the historical best solution's fitness value as convergence to the global optimum. They only proved the sequence converges, yet claimed PSO converges to the global optimum with probability 1. Moreover, the proof we provide later will demonstrate that the convergence of the fitness sequence and the convergence to global optimum are actually mutually exclusive properties of algorithms, and thus misuse could lead to undesirable results. Therefore, we believe it is necessary to classify these convergence studies. In view of this, we divide the studies of convergence as stable convergence analysis and global convergence analysis in this article.

Researches on stable convergence mainly focus on investigating whether the evolution trajectory of the optimal individual or the final positions of all individuals converge within a small region or to a certain point under infinite time. It should be noted that the region/point here does not have to cover/be the global optimum or even the local optimum. In detail, research on stable convergence analysis can be broadly categorized into the analysis of individual convergence [24]-[31] and the analysis of population convergence [32]-[34]. Individual convergence primarily aims to describe whether the trajectories of individuals in the algorithm stagnate, while population convergence generally aims to describe whether the population density or distribution tends to concentrate in a specific region. Furthermore, the analysis of individual convergence can be divided, akin to stability in control systems, into output stability analysis in classical control theory with different orders [24]-[29] and the Lyapunov stability analysis [29][30]; the analysis of population convergence mainly focus on the population distribution analysis [31]-[33]. For scenarios requiring a proof of **stable convergence**, the analysis might expect the algorithm to exhibit high robustness, ensuring that results do not vary significantly within a given time frame. This stability allows us to reliably estimate the algorithm's performance.

For scenarios requiring a proof of **global convergence**, the analysis prioritizes the algorithm's ability to escape local optima. Most of the existing researches on global convergence primarily focus on whether an algorithm can converge to the global optimum with probability 1 under infinite time. Specifically, the methods examine whether the historical best individual can be guaranteed to converge to the global optimum or an acceptable neighborhood around the global optimum. For the analysis of stochastic algorithms, Solis and Wets [35] proposed two classic assumptions to guarantee global optimality. Subsequently, Rudolph [35] applied these assumptions to EC algorithms for global convergence analysis by modeling the EC algorithms as homogeneous Markov chains. Currently, the mainstream researches [37]-[41] about global convergence analysis for EC algorithms are primarily based on the method given in [35]. Fig. 1 shows the taxonomy for research in theoretical analysis.

However, the method in [35] has the following two issues: On one hand, some advanced EC algorithms adaptively control the parameters or the search strategies based on the current population state, making them unsuitable for being modeled as homogeneous Markov chains, which might lead to misuse and consulting in improper conclusion [41]-[43]. On the other hand, even if the EC algorithms can be modeled as homogeneous Markov chains, the method in [36] imposes stricter assumptions than that of [35], excluding some algorithms that could otherwise ensure global convergence. Fig. 2 is a Venn diagram showing the relationship between the global convergence EC algorithms identified by the method in [36] and the true global convergence EC algorithms.

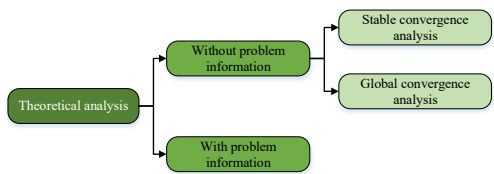

Fig. 1. Taxonomy for the research in theoretical analysis.

For example, in our observations, some studies [41][42] analyze the grey wolf algorithm using the method in [36] by mistakenly treating the time-varying parameter as non-time-varying parameter. By excluding time-varying parameter from the state, they modeled the algorithm as a homogeneous Markov chain, rendering their analysis of global convergence and final conclusions questionable. A similar yet not identical example is found in the analysis of the basic ant colony algorithm [43], where the authors only considered a part of the whole state for the Markov chain in the proof and thus leading the proof become questionable (although the final conclusion is correct). The above examples show that when using the method in [35] for global convergence analysis, users might easily mis-model algorithms due to incomplete understanding, leading to a questionable conclusion.

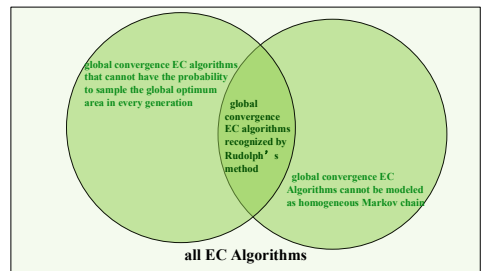

Fig. 2. Venn diagram of global convergence EC algorithms identified by Rudolph's method in [36] and actual global convergence EC algorithms.

Therefore, to assist better analyzing global convergence more effectively, we propose a new scope and domain measure comparison-based (SDMC) method for global convergence analysis. In SDMC, we propose a hypothesis **(H5)** to replace the **(H2)** in [35] and provide a rigorous proof that this **(H5)** is a necessary and sufficient condition for algorithms satisfying **(H1)** in [35] to guarantee global convergence. The SDMC method allows us to analyze an algorithm's global convergence without focusing on the sampling probability of arbitrary subsets, instead directing our attention to the more accessible metric, i.e., the measure of the algorithm's search scope.

In the SDMC method, we use the term *search Scope* to refer to the set of positions that all individuals of the algorithm have probability to reach at generation $t$, while the *feasible Domain* is the feasible region of the problem, independent of the algorithm. In our problem context, the *Measures* of these scopes and domains can be understood as their volumes in high-dimensional space. We encourage researchers to employ various reasonable methods to estimate the above-mentioned measures, and in the subsequent examples which we use to demonstrate the SDMC method, we primarily utilize tools from stable convergence analysis. We determine whether an algorithm satisfies **(H5)** as follow: if, for each time $t$, we make *Comparisons* between the measure of union $U_{t,N}$ of the algorithm's search scopes over a finite number $N$ generations from $t$ and the measure of the problem's feasible domain. If there exist an $N$ such that the above two measures are equal, then the algorithm guarantees global convergence; otherwise, it does not. We hope this work can inspire researchers to use some well-established theoretical tools from other fields to help simplify the analysis of global convergence.

The novelty and advantage of the SDMC method mainly lie in three aspects. First, it does not need to model the algorithms as homogeneous Markov chains, being applicable to algorithms that are not suitable for modeling homogeneous Markov chains such as those algorithms with time-variant parameters or settings. Second, it avoids tightening the hypothesis for convergence in [36]. Third, it is relatively simple and easy to understand, requiring only basic knowledge of probability theory and linear algebra, thus avoid a significant portion of misuse and concept confusion during the analysis.

To explore how theoretical analysis can do help to the design of algorithms, we further apply the proposed SDMC method to evaluate the impact of the gene targeting (GT) technology on the global convergence in a series of GT-based algorithms, GT-based differential evolution (GTDE) [45] and GT-based PSO (GTPSO) [46], and their original algorithms. Based on this, we offer suggestions on designing more efficient algorithms that ensure global convergence.

The contributions of this article are as follows:

Firstly, we give detailed descriptions of stable convergence and global convergence and prove a theorem demonstrating the mutual exclusivity of these two types of convergence. Furthermore, we illustrate the drawbacks of using the homogeneous Markov chain for global convergence analysis on EC algorithms, which inspires us to use other methods to analysis global convergence.

Secondly, we propose the new SDMC method for global convergence analysis. The proposed SDMC method does not rely on modeling the algorithm as homogeneous Markov chain, thus avoiding the tightening of hypothesis from[34] to [35] and simplifying the analysis. We analyze linear decreasing inertia weight PSO (LDIW-PSO) [46] and a very simple periodic partitioned random sampling as examples to demonstrate how the SDMC method can be applied to algorithms for which the method in [35] is unsuitable.

Thirdly, taking GTDE and GTPSO as examples, we analyze the role of the GT strategy for large-scale optimization. We discuss why GT can improve DE and social learning PSO (SLPSO) and achieve better results. Afterwards, we conclude under what circumstances GT can perform better and propose some suggestions for transplanting GT and designing new GT integrating algorithms which can guarantee global convergence. Accordingly, a very simple improvement for GTPSO is proposed and it obtains better performance.

The structure of this article is organized as follows. Section II describes the stable convergence and global convergence, where we provide the mutually exclusive relationship between them, which inspires the proposing of our method. Section III discuss the drawbacks of current commonly used analysis methods, followed by our proposed novel SDMC method. Section IV gives examples to show how to analyze the global convergence of algorithms via the proposed SDMC method. Section V analyzes two groups of GT-based algorithms and their original counterparts, exploring the role of GT in enhancing global convergence and thus we provide several suggestions on how to use GT and design globally convergent algorithms. Building on this, we propose an improvement to GTPSO, with experiments validating its feasibility. Section VI summarizes our work and offers prospects for future research.

## II. STABLE CONVERGENCE AND GLOBAL CONVERGENCE

### A. Stably Convergence

In the analysis of individual convergence, the *convergence* concepts from control systems or random variables are often employed. Overall, the goal is to describe whether the positions of individuals in the algorithm tend to stabilize. In other words, it evaluates whether the algorithm can gradually converge to a certain region or point over time. Here, we introduce different types of stability based on the stability classifications in control systems, which can primarily be divided into output stability and Lyapunov stability. Fig. 3 gives a taxonomy for the researches in stable convergence analysis.

We characterize the stable convergence into individual convergence analysis [23]-[30] and population convergence analysis [31]-[33]. The mainstream method for individual convergence analysis, which we classify them as output stability analysis, generally treats the individuals as second-order (or higher-order) systems and analyzes the stability of these systems. The time and frequency domain analysis of the transfer function is carried out using classical methods such as the algebraic criterion (Laws-Hallwitz criterion) and root locus

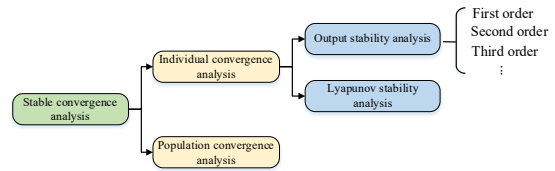

Fig. 3. Taxonomy for the research in stable convergence analysis.

criterion.

In the existing analysis, the definition of $M^{\text{th}}$-order stable convergence is as follows:
$$\lim_{t \to +\infty} E[x(t)^M] - P^M = 0 \quad (1)$$
where $x(t)$ is the position of the individual at time $t$ and $P$ is a constant that typically depends on both the algorithm and the problem.

The analysis of the algorithm's first-order convergence is generally found in [23]-[26], with a few works focusing on the second-order [27] and third-order [28] convergence of the algorithm.

However, the operators in EC algorithms (such as mutation and crossover) introduce a significant amount of randomness, which makes their behavior stochastic and uncertain. Thus, classical methods are no longer applicable for the convergence analysis of such EC algorithms. As a result, some studies, which we classify as Lyapunov stability analysis, directly use Lyapunov methods to analyze individuals, allowing for the determination of convergence in systems with random disturbances [29][30]. The Lyapunov method is based on the state space description method of the system, which is a general method that applies to the stability analysis for single variable, multiple variables, linear, nonlinear, constant, and time-varying systems. It not only describes the external characteristics of the system, but also reveals its internal characteristics.

In the analysis of population distribution, Wang and Huang [32] assume that all individuals in the population are independent and identically distributed, and use the probability density function (PDF) of every individual's distribution to analyze the population's evolutionary process. It demonstrates how the whole population's PDF changes through different operators. Although we believe that the entire population cannot be considered independent and identically distributed on many algorithm variants, this assumption greatly simplifies the analysis and also explains the overall behavior of algorithm evolution to a certain extent. This problem can be well handled in some simple situation, for example, we will classify and discuss different particles in the subsequence analysis of DE.

### B. Global Convergence

Global convergence is defined as whether the best individual in every generation guarantees to converge to the global optimum. The widely used definition of the global converge algorithm is proposed by Solis and Wets [34], where the sufficient conditions for the general convergence of random search algorithms are as follows:

Considering a measurable function $f: \mathbb{R}^n \to \mathbb{R}, S \subseteq \mathbb{R}^n$ and $S$ is a measurable set. To optimize the function $f$, we need to find an $x^* \in S$ that makes $f(x^*)$ to be an acceptable approximation of the infimum of $f$.

To find such $x^*$, the process of an optimization algorithm can be summarized as:

*Step 1*: Set $t = 0$, initialize $\xi^0$ and obtain $x^0$ from $\xi^0$;

*Step 2*: Generate $\xi^t$ from the sample space $(\mathbb{R}^n, \mathcal{B}, \mu_t)$;

*Step 3*: Set $x^{t+1} = D(x^t, \xi^t)$, choose $\mu_{t+1}$, set $t = t + 1$, and return to *Step 1* until $x^t$ becomes an acceptable solution.

In the above process, $\xi^t \in \mathbb{R}^n$ follows the distribution $\mu_t$, which is the conditional probability measures corresponding to distribution functions defined on $\mathbb{R}^n$; $\mathcal{B}$ can be any Borel subset of $\mathbb{R}^n$; $D(\cdot)$ is a map decided by the algorithm that $x^{t+1}$ can be obtained by $D(x^t, \xi^t)$.

Then Solis and Wets proposed the **THEOREM 1**:

**THEOREM 1:** For the above optimization algorithm, two hypothesizes are needed to support its convergence condition:

**(H1)** $f(D(x,\xi)) \leq f(x)$ and if $x \in S$, $f(D(x,\xi)) \leq f(\xi)$.

**(H2)** For any (Borel) subset $A$ of $S$ with $v(A) > 0$, there has $\prod_{t=0}^{\infty}[1 - \mu^t(A)] = 0$, in which $v$ is a nonnegative measure defined on $\mathcal{B}$ with $v(S) > 0$ and $\mu^t(A)$ is the probability that $A$ was produced by $\mu^t$.

The particularly similar forms between the $1 - \mu_t(A)$ discussed here and the widely studied convergence rate $1 - \pi_t(X^*)$ in the runtime analysis [47][48] may naturally lead to confusion of these two concepts, but they are completely different. Here, $\mu_t(A)$ merely denotes the probability that set $A$ is sampled at time $t$, while $\pi_t(X^*)$ refers to the probability that $X^*$ has been found before time $t$. Thus, $\pi_t(X^*)$ in the convergence rate is actually $\sum_{t'=1}^{t} \mu_{t'}(X^*)$.

The most famous transformation of the above definition to the field of EC can be found in [35] by Rudolph. That study models the algorithms as homogeneous Markov chains with the $t$-th generation of the Markovian kernel as:

$$K^{(t)}(x, A) = \begin{cases} K(x, A), & t = 1 \\ \int_E K^{(t)}(y, A) K(x, dy), & t > 1 \end{cases} \quad (2)$$

where $K(x, A) = P\{x^{t+1} \in A \mid x^t = x\}$. Since the algorithm is modeled as a homogeneous Markov chain, $K(x, A)$ will not change over time. Thus Rudolph [35] gives **THEOREM 2** as the global convergence definition for EC algorithms, as:

**THEOREM 2:** An EC algorithm that satisfies the following two hypotheses will converge to the global optimum of a real-value function $f: \mathbb{R}^n \to \mathbb{R}$ defined on an arbitrary space:

**(H3)** $K(x, A_\varepsilon) > \delta > 0$ for all $x \in A_\varepsilon^c = S \setminus A_\varepsilon$.

**(H4)** $K(x, A_\varepsilon) = 1$ for $x \in A_\varepsilon$.

where $A_\varepsilon = \{x \in S: d(x) < \varepsilon\}$ with some $\varepsilon > 0$ be the set of $\varepsilon$-optimal states and $d(x) = |f(x) - f(x^*)|$.

The proof of **THEOREM 2** is given in [35] and also briefly described as follows.

*Proof*: For $t \geq 1$ we have:

$$\begin{aligned} &K^{(t+1)}(x, A_\varepsilon) \\ &= \int_S K^{(t)}(y, A_\varepsilon) K(x, dy) \\ &= \int_{A_\varepsilon} K^{(t)}(y, A_\varepsilon) K(x, dy) + \int_{A_\varepsilon^c} K^{(t)}(y, A_\varepsilon) K(x, dy) \\ &= K(x, A_\varepsilon) + \int_{A_\varepsilon^c} K^{(t)}(y, A_\varepsilon) K(x, dy) \\ &\geq K(x, A_\varepsilon) + [1-(1-\delta)^t] \int_{A_\varepsilon^c} K(x, dy) \\ &= 1 - (1-\delta)^t (1 - K(x, A_\varepsilon)) \\ &\geq 1 - (1-\delta)^t (1-\delta) \\ &= 1 - (1-\delta)^{t+1} \end{aligned} \quad (3)$$

Thus, we have $P\{x^t \in A_\varepsilon\} = 1 - (1-\delta)^t$, which is the probability that the Markov chain is in set $A_\varepsilon$ at step $t$. Therefore, we have:

$$P\{d(x^t) > \varepsilon\} = 1 - P\{x^t \in A_\varepsilon\} \leq (1-\delta)^t \to 0$$

Since $t \to +\infty$, $d(x^t)$ converges to 0 in probability. Meanwhile, as $\sum_{t=1}^{\infty} P\{d(x^t) > \varepsilon\} \leq \sum_{t=1}^{\infty} (1-\delta)^t = (1-\delta)/\delta < \infty$, $d(x^t)$ can converge completely to 0.

### C. Relation Between Stable and Global Convergence

For the stable convergence and global convergence, we find that they in fact hold two kinds of mutually exclusive properties. Herein, we show the mutual exclusion of stable and global convergence in **THEOREM 3** to also further explain why we can use the tools in stable convergence analysis to analyze global convergence.

**THEOREM 3**: An algorithm that ensures stable convergence for all individuals cannot guarantee global convergence.

*Proof*: According to the definition of stable convergence of an algorithm, it must have the following:

$$\lim_{t \to +\infty} |\Delta x_i(t)| = \delta \quad (4)$$

where $\Delta x_i(t)$ is the position change of individual $i$ from $t$-1$^{\text{th}}$ generation to $t^{\text{th}}$ generation, $\delta$ is a relatively small constant that holds any definition for system stability convergence that we mentioned above. Even for the most relaxed form of stability, any change in individual position will always return to a position within $\delta$ from the equilibrium point while the system is considered stable. We also have $\delta \leq v(S^d)$, where $v(S^d)$ is the measure of $S$ in $d^{\text{th}}$ dimension of variables.

For each individual $i$, the search scope $C_i(t)$ in the $t^{\text{th}}$ generation is a *DIM*-dimensional cube with each dimension within its own range. For the entire algorithm, the search scope (i.e., the bounded support) of $\mu_t$ is $M(t) = \{C_1(t) \cup C_2(t) \cup ... \cup C_N(t)\}$.

Therefore, we must have:

$$v(M(t)) \leq 2 \frac{\delta^{DIM} \Gamma^{DIM}(1/2)}{DIM \times \Gamma(DIM/2)} \quad (5)$$

where $t \to +\infty$, there must have $v(M(t)) < v(S)$ after a certain generation. Thus, it is obvious that **(H2)** must not be satisfied when all the individuals from algorithms ensure stability and the algorithm cannot guarantee global convergence. It is obvious in the above proof that the stable convergence of the population search scope will also fail to guarantee global convergence.

From the above conclusion, we can find that stable convergence and global convergence are two mutually exclusive properties that must be carefully distinguished when mentioned. This is intuitively reasonable, as an algorithm that is guaranteed to find the global optimum will not stagnate indefinitely at a fixed value or a small range during the evolution.

## III. SDMC Method

### A. The Drawback of Modeling Algorithms as Homogeneous Markov Chain

It is easy to find that the proof in Section II-B turns $\prod_{t=0}^{\infty}[1-\mu^t(A)] = 0$ in **Theorem 1** into a stricter hypothesis $K(x, A_\varepsilon) > \delta > 0$ to hold the two inequations.

However, we find that if $K(x, A)$ changes over time, then it will not need to always be positive to hold Eq. (3). Here we rewrite $K(x, A)$ as $K_t(x, A)$, which rely on the change of time and condition $K_t(x, A) \geq \delta_t \geq 0$, then Eq. (3) will be written as Eq. (6):

$$\begin{aligned}
&K^{(t+1)}(x, A_\varepsilon) \\
&= \int_S K^{(t)}(y, A_\varepsilon) K_t(x, dy) \\
&= \int_{A_\varepsilon} K^{(t)}(y, A_\varepsilon) K_t(x, dy) + \int_{A_\varepsilon^c} K^{(t)}(y, A_\varepsilon) K_t(x, dy) \\
&= \int_{A_\varepsilon} K_t(x, dy) + \int_{A_\varepsilon^c} K^{(t)}(y, A_\varepsilon) K_t(x, dy) \\
&\geq K(x, A_\varepsilon) + [1 - \prod_{t'=1}^{t}(1-\delta_{t'})] \int_{A_\varepsilon^c} K_t(x, dy) \\
&= 1 - (1-\delta)^t (1 - K_t(x, A_\varepsilon)) \\
&\geq 1 - \prod_{t'=1}^{t+1}(1-\delta_{t'})
\end{aligned} \quad (6)$$

We can find from Eq. (6) that it only takes the probability of sampling any subset $A_\varepsilon$ that does not remain 0 indefinitely, then $d(x^t)$ can converge to 0 in probability.

That is, under the infinite generations, any subset $A_\varepsilon$ should have infinite generations that provide it the probability of being sampled, instead of having the probability of being sampled in every generation, a.k.a. **(H3)**.

Thus, we identify another drawback of modeling EC algorithms as homogeneous Markov chains. This method inherently tightens the hypothesis for global convergence, leading to some algorithms that are inherently globally convergent being incorrectly deemed incapable of guaranteeing global convergence. Moreover, it is not suitable for more advanced algorithm variants where parameters change over time.

### B. The Scope and Domain Measure Comparison-based (SDMC) Method

In Section III-B, we have demonstrated how stability and global convergence become two mutually exclusive properties. From the proof of **Theorem 3**, we can observe that when individual trajectories converge, there is always a corresponding convergence of the population's search scope, which leads to the conclusion that global convergence is not satisfied. Additionally, since algorithms typically have global boundary-handling mechanism, we can determine whether an algorithm satisfies global convergence by comparing the measures between the population's search scope and the feasible domain.

Thus, we give a **(H5)** instead of **(H2)** for EC algorithms to obtain the SDMC method, who only need to concern about the measure of the algorithm's search scope in some generations and the feasible domain, without considering any possible subset of the whole feasible domain:

**(H5):** $\forall t \neq +\infty$, $\exists N \neq +\infty$ that $v(U_{t,N}) = v(S)$, in which $U_{t,N} = \bigcup_{k=0}^{N-1} M(t+k)$.

We conclude that **(H5)** is the sufficient and necessary condition for algorithm who satisfied **(H1)** to be a convergence algorithm. Here we give the proof.

***Proof***: We first simplify the problem as $\{M(t)\}_{t=1}^{+\infty}$ to be a sequence of subsets of $S$, which is reasonable with the help of global boundary handling mechanism so that $M(t) \subseteq S$ holds for all $t$. Therefore, we have $v(U_{t,N}) = v(S)$ equal to $U_{t,N} = S$.

For any $A \subseteq S$ with $v(A) > 0$, $G = \mathbb{N}$ to be the set of generations, $T(A) = \{t \in G \mid M(t) \cap A \neq \emptyset\}$ is the set of generations that $A$ has the probability to be sampled, and $|T(A)|$ is the number of these generations. Let $\mu_t(A)$ represent $A$ will be sampled at generation $t$, the probability that $A$ is never sampled over infinite time is given by:

$$P_{\bar{A}} = P(\bigcap_{t=0}^{\infty} \{x^t \notin A\}) = \prod_{t=0}^{\infty}(1 - \mu_t(A)) \quad (7)$$

Here we first consider the efficiency of **(H5)**, that is, we will have Eq. (7) = 0 when **(H5)** holds.

We partition all the generations ($G = \mathbb{N}$) into infinitely many intervals of length $N$: $I_j = [jN, (j+1)N-1]$, where $j \in \mathbb{N}$. For each $j$, we define $U_{jN,N} = \bigcup_{k=0}^{N-1} M(jN+k)$, and $U_{jN,N} = S$ when **(H5)** holds.

Therefore, we have $\forall j, \forall A, A \cap U_{jN,N} \neq \emptyset$, thus there must exist $k_j \in \{0, 1, ..., N-1\}$ such that $M(M(jN+k_j)) \cap \neq \emptyset$.

We then define $t_j = jN + k_j$, then we have: $\forall j, t_j \in T(A)$.

Since $t_{j+1} = (j+1)N + k_{j+1} \geq (j+1)N > jN + N - 1 \geq jN + k_j = t_j$, the sequence $t_j$ is strictly increasing and thus $\{t_j \mid j \in N\}$ is an infinite subset of $T(A)$, thus we also have $|T(A)| = +\infty$, leading to a result that Eq. (7) = 0 holds as $0 \leq \mu_t(A) \leq 1$ for all $t$ and it is the product of infinitely many values less than 1.

In a word, the probability of subset $A$ being sampled is 1, which ensures the efficiency of **(H5)**.

For the necessity, we first assume that **(H5)** does not hold, i.e., $\exists t_s^N \neq +\infty$, $U_{t_s^N,N} = \bigcup_{k=0}^{N-1} M(t_s^N + k) \neq S$ for all $N \in G$. Let $A_s = S \setminus U_{t_s,N}$ and it is obvious that $A_s \neq \emptyset$.

As $t_s^N \neq +\infty$, we can easily have $T(A_s) < t_s^N < +\infty$. Here we set $\mu_{max} = \sup\{\mu_t(A) \mid t \leq t_s^N\}$, we have $P_{\bar{A}} > (1-\mu_{max})^{t_s^N} = \sigma$, where $\sigma$ is a maybe very small but non zero value.

It is obvious that when **(H5)** is not satisfied, we can always construct a counterexample $A_s$ who cannot be sampled with probability 1.

Thus, we finish the proof that **(H5)** is a necessary and sufficient condition for an EC algorithm satisfying **(H1)** to guarantee global convergence.

The SDMC method is not only more general (in analyzing algorithms that cannot be modeled as homogeneous Markov chains) and more accurate (in identifying algorithms where the sampling in every generation does not cover the entire feasible domain) compared to the method proposed in [35] given by Rudolph, but also simpler than the method given by Solis and Wets [35]. The key advantage lies in the fact that we only need to concern about the search scope of the algorithms instead of the sampling probability of any possible subset of the feasible domain. Furthermore, although our criterion appears to involve $N$, we do not actually need to determine its exact value, but only to make sure that it is not positive infinity. Our method has advantage in analyzing algorithms with mechanisms that can adaptively adjust based on evolutionary states, such as reset [49]-[51] and reinitialize[52][53].

## IV. Validation of SDMC for Global Analysis

As mentioned above, our SDMC method is more practical when analyzing algorithms that cannot be modeled as homogeneous Markov chains and algorithms that cannot satisfy the probability of sampling the entire feasible domain in every generation.

Herein, we select an example from each of these two types of algorithms to demonstrate the practicality of the SDMC method.

*A. Analysis for Algorithms that Cannot be Modeled as Homogeneous Markov Chain*

Herein, we take LDIW-PSO [46] as an example to show how to analysis algorithms that cannot be modeled as homogeneous Markov chain with the SDMC method.

We first explain why this algorithm cannot be analysis with the method in [35].

The position of particle $i$ in generation $t$ of LDIW-PSO is updated as follow:

$$x_i(t+1) = x_i(t) + v_i(t+1) \quad (8)$$

$$v_i(t+1) = \omega(t)v_i(t) + c_1 r_1 (pbest_i(t) - x_i(t)) \\ + c_2 r_2 (gbest_i(t) - x_i(t)) \quad (9)$$

in which $c_1$ and $c_2$ are preset parameters (The parameter settings are $c_1 = 2$ and $c_2 = 2$ in [46]), $r_1$ and $r_2$ are random numbers between [0, 1] generated from uniform distribution in every generation. $\omega(t)$ is updated as follow:

$$\omega(t) = \omega_{start}(\omega_{start} - \omega_{end})\frac{T_{max} - t}{T_{max}} \quad (10)$$

in which $\omega_{start} = 0.9$ and $\omega_{end} = 0.4$ are set respectively in [46].

Thus, for the particle $i$ in dimension $d$, we set $\Delta_1 = c_1 r_1(t)(pbest_{i,d}(t) - x_{i,d}(t))$, $\Delta_2 = c_2 r_2(t)(gbest_d(t) - x_{i,d}(t))$, $\delta_1 = c_1(pbest_{i,d}(t) - x_{i,d}(t)) \cap c_2(gbest_d(t) - x_{i,d}(t))$, $\delta_2 = c_1 (pbest_{i,d}(t) - x_{i,d}(t)) \cup c_2 (gbest_d(t) - x_{i,d}(t))$. Then we have $Z = v_{max} - \omega(t)v_{i,d}(t)$. We can find that given $\Delta_1$ and $\Delta_2$, $Z$ is the parallelogram shown in Fig. 4, with vertices at $(0, 0, 0)$, $(\delta_1, 0, \delta_1)$, $(0, \delta_2, \delta_2)$, and $(\delta_1, \delta_2, \delta_1 + \delta_2)$.

It is easy to obtain that the probability density of $Z$, is the ratio of the length of the line segment of the parallelogram on the plane corresponding to the value of $Z$ to the total area of the parallelogram, i.e.:

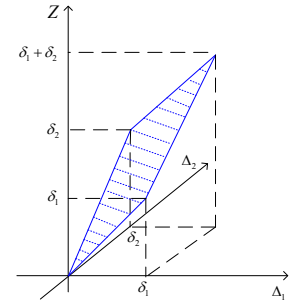

Fig. 4. The value of $Z$ given $\Delta_1$ and $\Delta_2$ for LDIW-PSO.

$$f_z(\delta_1 + \delta_2 < Z) = \begin{cases} \frac{\sqrt{2}Z}{\sqrt{3}\delta_1\delta_2}, & 0 \leq Z \leq \delta_1 \\ \frac{\sqrt{2}\delta_1}{\sqrt{3}\delta_1\delta_2}, & \delta_1 \leq Z \leq \delta_2 \\ \frac{\sqrt{2}(Z-\delta_1)}{\sqrt{3}\delta_1\delta_2}, & \delta_2 \leq Z \leq \delta_1 + \delta_2 \\ 0, & \text{otherwise} \end{cases}$$

Consider about the $v_{max}$ as the velocity limitation, we have:

$$P(v_{max} = v_{i,d}(t+1)) = \int_0^{v_{max}} f_z(\delta_1 + \delta_2 < Z)dZ$$

Then we have the transition probability for the position of particle (without considering the feasible domain):

$$P(x_{i,d}(t) \to x_{i,d}(t+1)) = \begin{cases} 1 - \int_0^{v_{max}} f_z(\delta_1 + \delta_2 < Z)dZ, & x_{i,d}(t+1) = x_{i,d}(t) + v_{max} \\ f_z(\delta_1 + \delta_2 < Z), & x_{i,d}(t+1) \in [x_{i,d}(t) + \omega(t)v_{i,d}(t), x_{i,d}(t) + v_{max}] \\ 0, & \text{otherwise} \end{cases}$$

It is evident that the transition probability is related to $\omega(t)$, thus the transition probability from any state $j$ to another state $k$ is also related to $\omega(t)$. We set $S_i(t) = (c_1, c_2, gbest, x_i, pbest_i)$ as the state of individual $i$ in generation $t$, then its obvious that $P(S_j(0) \mid S_k(0)) \neq P(S_j(t') \mid S_k(t'))$, where $t'$ is an arbitary given time.

In a word, LDIW-PSO is not homogeneous thus fails to meet the requirements for the Rudolph's method [35], let alone applying this method for analysis.

Here we will show how we analysis with the SDMC method. Let's correspond the variables in LDIW-PSO to the variables in Section II-B. If the state for time $t$ (the solution sequence $x^t$ in Section II-B) could be $gbest(t)$ in LDIW-PSO, thus we set the $X(t)=\{pbest_1(t), pbest_2(t), \ldots, pbest_N(t)\}$ to be $\xi^t$ so the function $D(\cdot)$ would set as:

$$D(gbest(t-1), X(t)) = \begin{cases} gbest(t), & if\ f(gbest(t)) < f(gbest(t-1)) \\ gbest(t-1) & otherwise \end{cases} \quad (11)$$

where we also have:
$$f(gbest(t)) = \min(f(pbest_1(t), f(pbest_2(t), \ldots, f(pbest_N(t))))$$

in the $t^{th}$ generation.

We consider the 'best' case first in which for each particle $i$, the *pbest* could always be updated. Then we have:
$$pbest_{i,d}(t) = x_{i,d}(t) \quad (12)$$

We model particles in LDIW-PSO as a second-order linear system, like what [23]-[26] do to analysis its stable convergence:

$$\begin{bmatrix} x_i(t+1) \\ v_i(t+1) \end{bmatrix} = \begin{bmatrix} 1-(c_1r_1+c_2r_2) & \omega(t) \\ -(c_1r_1+c_2r_2) & \omega(t) \end{bmatrix} \begin{bmatrix} x_i(t) \\ v_i(t) \end{bmatrix} \quad (13)$$
$$+ c_1r_1 pbest_i(t) + c_2r_2 gbest(t)$$

We set $A$ as the state matrix, whose eigenvalues can help us to analyze the stable convergence for particle $i$. Here we have the eigenvalues for $A$ as:

$$\lambda_{1,2} = \frac{1+\omega \pm \sqrt{(1+\omega)^2 - 4[\omega - (c_1r_1+c_2r_2)]}}{2} \quad (14)$$

According to the parameter setting of LDIW-PSO, we have the expectation for the maximum module of eigenvalues of $A$ as:

$$E_{start}[\max(|\lambda_1|,|\lambda_1|)] = \max(|\frac{0.9 \pm 1.64317i}{2}|) \approx 0.94854$$

$$E_{end}[\max(|\lambda_1|,|\lambda_1|)] = \max(|\frac{0.4 \pm 1.78885i}{2}|) \approx 0.84$$

We can easily obtain the two value is the maximum and the minimum expectation of the maximum module of eigenvalues in the whole evolutionary process of the LDIW-PSO when ω keeps decreasing. Then we conclude that every particle in LDIW-PSO is stable convergence in probability, that is, $\lim_{t\to\infty} E(\Delta x_i(t)) = 0$. Then we must have $\lim_{t\to\infty} v(M(t)) = 0$ as $v(M(t))$ keeps reducing, there must have a certain generation $T'$ that after which $U_{T',+\infty} \neq S$. Thus (H5) could not be satisfied.

In other cases that are not 'best', we naturally have $C_i(t)=0$ (when $f(x_i(t)) < f(pbest_i(t-1))$ no longer satisfied), leading to a smaller $v(M(t))$ than the 'best' case. Therefore, it is even less possible to satisfied **(H5)**.

Therefore, we conclude that LDIW-PSO cannot guarantee global convergence according to the analysis by using our SDMC method.

### B. Analysis for Global Convergence Algorithms Cannot Satisfy *(H3)*

Since we have not yet identified a widely used representative algorithm, we will describe a very simple counterexample in this section. This counterexample, even when modeled as a homogeneous Markov chain, guarantees global convergence without satisfying **(H3)**.

This method is a variant of random sampling. We arbitrarily divide the feasible domain $S$ into two subsets, $B$ and $C$, where $B = C \setminus S$ is the complement of $C$ in $S$. Our method alternates between performing random sampling obeying a uniform distribution on $B$ and $C$ in each generation. That is:

$$K(x, A_\varepsilon) = P\{X_t \in A_\varepsilon \mid X_{t-1}\} = \begin{cases} \frac{v(A_\varepsilon)}{v(B)}, & if\ A_\varepsilon \subseteq B\ and\ X_{t-1} \notin B \\ \frac{v(A_\varepsilon)}{v(C)}, & if\ A_\varepsilon \subseteq C\ and\ X_{t-1} \notin C \\ 0, & otherwise \end{cases} \quad (15)$$

It is obvious that **(H3)** cannot hold since $K(x, A_\varepsilon) = 0$ for $x$ belongs to the subset which was selected for sampling in the previous generation, leading to the conclusion that this sampling method cannot guarantee convergence to the global optimum.

However, conclusion will be different if we use SDMC to analysis this sampling method, for whom $\forall t \neq +\infty$, $N = 2$, $v(U_{t,N}) = v(S)$.

The conclusion obtained using the analysis method in [34] validates our conclusion. Because for any $A$, we have:

$$\prod_{t=0}^{\infty}[1-\mu^t(A)] = \begin{cases} \prod_{t=0}^{\infty/2}[1-\frac{v(A_\varepsilon)}{v(B)}], & if\ A_\varepsilon \subseteq B \\ \prod_{t=0}^{\infty/2}[1-\frac{v(A_\varepsilon)}{v(B)}] & if\ A_\varepsilon \subseteq C \end{cases}$$

which will obvious become 0 either $A_\varepsilon \subseteq B$ or $A_\varepsilon \subseteq C$.

## V. APPLICATION OF SDMC FOR ALGORITHM DESIGN

To explore the application role of theoretical analysis in algorithm design, we analyze two groups of GT-based algorithms and their original versions, differential evolution (DE) [54] and social learning PSO (SLPSO) [56] to determine the impact of GT on their global convergence. Our theoretical analyses results show that:

- DE does not guarantee to converge to global optimum, but the incorporation of GT helps GTDE enhance its global convergence. More specifically, under the original parameter settings described in the GTDE, it can guarantee global convergence;
- SLPSO does not guarantee to converge to global optimum, but we find that it has better global convergency than DE;
- GTPSO does not guarantee to converge to global optimum, but we find that GT can help to improve its global convergency comparing to the basic SLPSO.

Here follows the detail of our analysis.

### A. Analysis of DE and GTDE
#### 1) DE
DE, proposed by Storn and Price [54], is one of the most well-known evolutionary algorithms in the EC family, for its ease of use and effectiveness. The main operations in DE include mutation and crossover. There are various commonly used operators for DE mutation, such as DE/rand/1, DE/best/1, and DE/current-to-best/1. Herein, the DE/current-to-best/1 mutation operator is adopted, as:

$$v_i(t+1) = x_i(t) + F \times (x_{best}(t) - x_i(t)) + F(x_{r1}(t) - x_{r2}(t)) \quad (16)$$

where $r1$ and $r2$ are individuals randomly selected in the population and they must be different. The setting of parameter $F$ is also different from the basic DE that it becomes a random number generated by Gaussian distribution with a mean of 0.7 and a standard deviation of 0.5. If $v_i(t+1)$ is out of the feasible scope, a new solution will be randomly created within the feasible range of the solution space.

The crossover performed after the mutation is shown as:

$$u_{i,d}(t+1) = \begin{cases} v_{i,d}(t+1), & \text{if } rand(0,1) < CR \text{ or } d = d_{rand} \\ x_{i,d}(t), & \text{otherwise} \end{cases} \quad (17)$$

where $d \in \{1, ..., DIM\}$ is the dimension of variables, $CR$ is the crossover rate, and $d_{rand}$ is a random integer generated between $[1, DIM]$.

After crossover, DE will decide whether $x_i$ is updated by $u_i$ or not, by:

$$x_i(t+1) = \begin{cases} u_i(t+1), & \text{if } f(u_i(t+1)) < f(x_i(t)) \\ x_i(t), & \text{otherwise} \end{cases} \quad (18)$$

*2) GTDE*

GTDE [45] performs GT operation on the best individual in every generation to obtain better solutions. In this process, there is a randomly generated number follows a uniform distribution in [0,1] for each dimension, which will be compared with another random number $P_j$ following a Gaussian distribution with a mean of 0.1 and a standard deviation of 0.01 to determine whether the dimension is a bottleneck dimension.

For each bottleneck dimension, GTDE generates a random number that follows uniform distribution in [0,1] to compare with a hyperparameter $P_m$ to determine which GT strategy to use. If the random number is less than $P_m$, $v_i(t+1)$ is calculated by:

$$v_i(t+1) = x_{best}(t) + F(x_{r1}(t) - x_{rand}) \quad (19)$$

otherwise, $v_i(t+1)$ is calculated by:

$$v_i(t+1) = x_{best}(t) + F \times (x_{r1}(t) - x_{r2}(t)) \quad (20)$$

where $x_{rand}$ is a randomly generated solution and $r1$ and $r2$ are two randomly selected integers from $\{1, 2, ..., N\}$.

*3) Analysis of DE*

Let's correspond the variables in DE to the variables we used before. If the state for time $t$ (the solution sequence $x^t$) could be $x_{best}(t)$ in DE, thus we set the $X(t)=\{x_1(t), x_2(t), ..., x_N(t)\}$ to be $\xi^t$ so the function $D(\cdot)$ would set as Eq.(21):

$$D(x_{best}(t-1), X(t)) = \begin{cases} x_{best}(t), & \text{if } f(x_{best}(t)) < f(x_{best}(t-1)) \\ x_{best}(t-1) & \text{otherwise} \end{cases} \quad (21)$$

in which $best = \arg\min(f(x_1(t), f(x_2(t), ..., f(x_N(t))))$ in the $t^{th}$ generation. Then we have:

$$\begin{aligned} & f(D(x_{best}(t), X(t+1))) \\ & = \min\{f(x_{best}(t)), f(x_i(t)) \text{ for } i \in \{1,2,...,N\}\} \end{aligned} \quad (22)$$

It is obvious that the **(H1)** could be satisfied.

For simplicity, $v(A)$, the Lebesgue measure of $A$, could be set as the *DIM*-dimensional volume of $A$. In the basic DE, for each individual $i$, the search space in dimension $d$ is $[r_{min,d}(t), r_{max,d}(t)]$, in which we have

$$\begin{aligned} r_{min,d}(t+1) = \min\{&x_{i,d}(t), x_{i,d}(t) \\ &+ F_{max} \times (x_{best,d}(t) - x_{i,d}(t) + x_{r1,d}(t) - x_{r2,d}(t)), \\ &x_{i,d}(t) + F_{min} \times (x_{best,d}(t) - x_{i,d}(t) + x_{r1,d}(t) - x_{r2,d}(t))\} \end{aligned} \quad (23)$$

$$\begin{aligned} r_{max,d}(t+1) = \max\{&x_{i,d}(t), x_{i,d}(t) \\ &+ F_{max} \times (x_{best,d}(t) - x_{i,d}(t) + x_{r1,d}(t) - x_{r2,d}(t)), \\ &x_{i,d}(t) + F_{min} \times (x_{best,d}(t) - x_{i,d}(t) + x_{r1,d}(t) - x_{r2,d}(t))\} \end{aligned} \quad (24)$$

It can be imagined that for individual $i$, the search scope in the $t^{th}$ generation is a *DIM*-dimensional cube with each dimension within the aforementioned range, which we call $C_i(t)$. For the entire algorithm, the search scope (a.k.a. a. the bounded support) of $\mu_t$ is $M(t) = \{C_1(t) \cup C_2(t) \cup ... \cup C_N(t)\}$.

Here we consider the 'best' case that for every dimension $d$ the $x_{i,d}$ could be updated that is, $rand(0,1) < CR$ for all the dimensions and $f(u_i(t)) < f(x_i(t-1))$. We have:

$$x_{i,d}(t+1) = (1-F)x_{i,d}(t) + F \times (x_{best,d}(t) + x_{r1,d}(t) - x_{r2,d}(t)) \quad (25)$$

When the algorithm begins to converge, we consider about the $\{x_{best,d}(t)\}_{\infty}^{t=0}$ first:

$$x_{best,d}(t+1) = x_{best,d}(t) + F \times (x_{r1,d}(t) - x_{r2,d}(t)) \quad (26)$$

$$E(\Delta x_{best,d}(t+1)) = E(F \times (x_{r1,d}(t) - x_{r2,d}(t))) \quad (27)$$

As $r1$ and $r2$ are randomly selected from the whole population, we assume $Z_{best}(t) = x_{r1,d}(t) - x_{r2,d}(t)$, thus we have the cumulative distribution function (CDF) of it:

$$\begin{aligned} F_{CDF}(Z_{best}(t)) &= P(x_{r1,d}(t) < Z_{best}(t) + x_{r2,d}(t)) \\ &= \int_{M_{min,d}(t)}^{M_{max,d}(t)} \int_{M_{min,d}(t)}^{M_{max,d}(t)} f_j(x_{r1,d}(t), x_{r2,d}(t)) dx_{r1,d}(t) dx_{r2,d}(t) \end{aligned} \quad (28)$$

As the CDF depends on the distribution $\mu_t$, we can consider with $t = 0$ first:

$$F_{CDF}(Z_{best,d}(0)) = \begin{cases} \dfrac{Z_{best,d}(0) + M1_{max,d}(0)}{2 \times M1_{max,d}(0)}, & -M1_{max,d}(0) \le Z_{best,d}(0) \le M1_{max,d}(0) \\ 0 & Z_{best,d}(0) < -M1_{max,d}(0) \\ 1 & Z_{best,d}(0) > M1_{max,d}(0) \end{cases}$$

Since we have $M_{max,d}(0) = x_{max,d}$ and $M_{min,d}(0) = x_{min,d}$, then $M1_{max,d}(0) = (x_{max,d} - x_{min,d})$ and its probability density function (PDF) becomes:

$$\begin{aligned} PDF(Z_{best,d}(0)) &= \dfrac{dF_{CDF}(Z_{best,d}(0))}{dZ_{best,d}(0)} \\ &= \begin{cases} \dfrac{1}{2*M1_{max,d}(0)} & -M1_{max,d}(0) < Z_{best,d}(0) < M1_{max,d}(0) \\ 0 & \text{otherwise} \end{cases} \end{aligned} \quad (29)$$

Without considering the constraint for the value of solutions, Eq. (27) would become:

$$E(\Delta x_{best,d}(2)) = E(F) \times E(x_{r1,d}(t) - x_{r2,d}(t)))$$
$$= F \times \int_{-M1_{max,d}(0)}^{M1_{max,d}(0)} \frac{Z_{best,d}(0)}{2M1_{max,d}(0)} dZ_{best,d}(0) \quad (30)$$
$$= F \times \frac{M1_{max,d}(0)}{2}$$

Thus for the individual $i$, the expectation of its search scope in dimension $d$ when $t = 1$ will reduced to a quarter of that when $t = 0$, as in the basic DE, we have $F = 0.5$.

And since the distribution of the best individual at the next generation $t + 1$ is always the distribution of the difference of two random variables following the distribution $\mu_{t,d}$ of the entire population. As we know, in the 'best' case hypothesis, the $x_{best,d}(t + 1)$ will follow the distribution given by the convolution of $\mu_{t,d}$ and the distribution $\mu^z_{t,d}$, from which $Z_{best,d}(t)$ was generated, multiplied by the factor $F$.

Since we all know that the variation of the distribution obtained by subtracting two random variables of the same distribution is:

$$Var(X1 - X2) = Var(X1) + Var(X2) = 2\sigma \quad (31)$$

where $X1$ and $X2$ are two random variables, $\sigma$ is the standard deviation of their original distribution. Thus, the standard deviation of $F \times \mu^z_{t+1,d}$ would become:

$$Var(F \times \mu^z_{t+1,d}) = 2F^2 \sigma_{t,d} \quad (32)$$

where $\sigma_{t,d}$ is the standard deviation of $\mu_{t,d}$.

As this convolution keeps working in every $\mu_{t,d}$ and $\mu^z_{t,d}$, the shape of the distribution $\mu_{t,d}$ gradually sharpens from a low trapezoid under the limit of $[x_{min,d}, x_{max,d}]$. That is, the probability of falling in the middle of the distribution is getting higher and higher, so the limit of Eq. (25) becomes:

$$\lim_{t \to \infty} E(\Delta x_{best,d}(t+1)) = 0 \quad (33)$$

This indicates that the trajectory of the best individual tends to converge to a stable value.

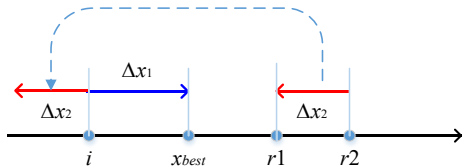

Fig. 5. Showing the evolution direction of individual $i$ of DE in dimension $d$.

For other individuals $i$, the search scope in dimension $d$ is shown in Fig. 5. We can assume that for individual $i$, the evolutionary direction can be regarded as the direction towards the best individual while adding a vector with a random direction and a modulus whose CDF is shown in Eq. (28).

Therefore, it is easy to obtain that for all individuals, their behavior tends to converge towards the best individual, leading to a result that $\lim_{t \to \infty} E(x_{r1,d}(t) - x_{r2,d}(t)) = 0$.

As we know the behavior of the entire population also tends to converge towards the best individual, we have $\lim_{t \to \infty} v(M(t)) = 0$ since in all dimension $r_{min,d}(t+1) = x_{i,d}(t)$ and $r_{max,d}(t+1) = x_{i,d}(t)$ when $t \to \infty$. Thus **(H5)** could not be satisfied as $v(M(t))$ keeps reducing, there must have a time $t'$ after which $v(U_{t',+\infty}) < v(S)$.

In other cases that are not 'best', we naturally have $C_i(t)=0$ (when $f(u_i(t)) < f(x_i(t-1))$ no longer satisfied) or $v(M_d(t)) = 0$ (when $rand(0,1) \geq CR$ for dimension $d$).

Therefore, DE cannot guarantee global convergence.

*4) Analysis of GTDE*

Return to GTDE, which provides two core improvements on the side of convergence. One is generating $F$ by a Gaussian distribution, and the other is the GT given by Eq. (19). Since $F$ is no longer limited to [0,1], when it is greater than 1, it partially makes reduction speed of $v(M(t))$ slower because $E(\Delta x_{best,d}(t+1))$ decreasing slower. The second is the GT in Eq. (19), which introduced the $x_{rand}$ thus Eq. (33) no longer holds and $\lim_{t \to \infty} v(M(t))$ becomes always positive.

Actually, **(H5)** is guaranteed to hold under the parameter setting given in [44], since $F$ follows a Gaussian distribution with a mean of 0.5 and a standard deviation of 0.1. Thus, it is possible to have $v(U_{t,N}) < v(S)$ for all $t$ with $N = 1$.

Even if the parameters were not set to guarantee global convergence, the improvements made by the GT strategy have successfully maintained a positive value of $v(M(t))$, resulting

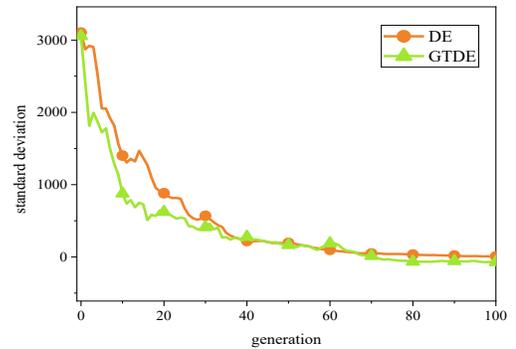

(a) On the early stage

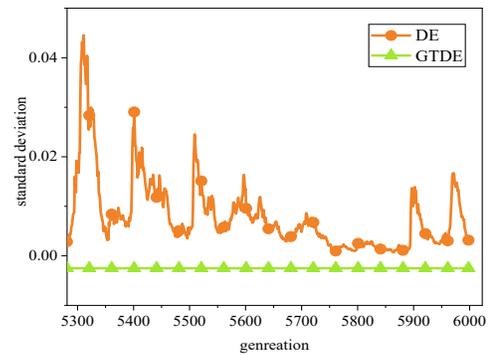

(b) On the later stage

Fig. 6. The standard deviation of the 1st dimension of the individuals in DE and GTDE on sphere function with DIM=1000 in different stages (a) On the early stage; (b) On the later stage.

in better diversity for the GTDE algorithm.

In other to visualize the improvements, we compare the standard deviation for DE and GTDE on sphere function at the 1st dimension, which should be a fair indicator for $v(M(t))$. The standard deviation of the 1st dimension of the individuals of two algorithms at the early stage and later stage are shown in Fig. 6 (a) and (b), respectively.

We can find in Fig. 6 that most of the time, the standard deviation of individuals in DE is always less than that of GTDE. At a later stage, when both algorithms converge (i.e., the standard deviation is very close to 0), GTDE still has the chance to jump out of this stagnant state and guide the particles to continue searching, which validates our conclusion above.

It is worth noting that since GTDE uses an additional number of adaptation calculations for GT in each generation, we compress the shape of DE here to ensure that the comparison in the standard deviation of the two algorithms at each time $t$ is with the same fitness evaluation.

### B. Analysis of SLPSO and GTPSO

#### 1) SLPSO

PSO [55], proposed by Kennedy and Eberhart in 1995, is one of the most representative swarm intelligence (SI) algorithms in the EC family, resulting in a lot of variants. SLPSO is a PSO variant proposed for large-scale optimization problems [56], in which the particles (i.e., solutions) within the current swarm (i.e., population) are sorted from best to worst based on their fitness values and a particle will learn from a randomly selected particle in all of the superior particles to update its position. For a particle $x_i$ at $t^{th}$ generation, its update process can be defined as:

$$v_{i,d}(t+1) = r_1 v_{i,d}(t) + r_2(x_{k,d}(t) - x_{i,d}(t)) + \varepsilon r_3(\overline{x_d}(t) - x_{i,d}(t)) \quad (34)$$

$$x_{i,d}(t+1) = \begin{cases} x_{i,d}(t) + v_{i,d}(t+1), & \text{if } rand(0,1) < P_i \\ x_{i,d}(t), & \text{otherwise} \end{cases} \quad (35)$$

where $r_1$, $r_2$, and $r_3$ are uniformly distributed random variables in [0,1], $k$ is the index of the randomly selected superior particle, and the $\varepsilon$ is proportional to the problem dimension and defined as:

$$\varepsilon = \beta \cdot \frac{DIM}{M} \quad (36)$$

Herein, a small value of $\beta = 0.01$ is used to avoid premature convergence in this work, $DIM$ is the dimension of the function and $M = 100$. $P_i$ is the learning probability for particle $i$ and is calculated as:

$$P_i = (1 - \frac{i-1}{N})^{\mu \cdot \log(\lceil \frac{DIM}{M} \rceil)} \quad (37)$$

in which $N$ is the population size and is set as $N = M + \lfloor 0.1 DIM \rfloor$; $\mu$ is set as 0.5.

#### 2) GTPSO

GT in GTPSO [45] is very similar to that in GTDE as they only act on the current best solution. The difference between GTDE and GTPSO is that GTPSO uses Eq. (38)-(39) to replace Eq. (19)-(20) in GTDE:

$$v_{i,d}(t+1) = \omega v_{i,d}(t) + c_1 r_1 (pbest_{k_1,d}(t) - pbest_{k_2,d}(t)) + c_2 r_2 (\overline{x_d}(t) - x_{best,d}(t)) \quad (38)$$

$$v_{i,d}(t+1) = Gaussian(\frac{1}{2}(v_{k_1,d}(t) + v_{k_2,d}(t)), \frac{1}{2}(v_{k_1,d}(t) - v_{k_2,d}(t))) \quad (39)$$

where $\omega=0.4$, $k_1$ and $k_2$ are the index of two randomly selected particles and $\overline{x_d}$ is the mean position of all the particles in $d^{th}$ dimension.

#### 3) Analysis of SLPSO

For various variants of PSO, $x_{best}(t)$ is no longer suitable as the state of time $t$, otherwise **(H1)** will never be satisfied. We will use $gbest(t)$ instead thus **(H1)** can hold.

We set the $P(t) = \{pbest_1(t), pbest_2(t), \ldots, pbest_N(t)\}$ to be $\xi^t$ and the function $D(\cdot)$ is set as Eq.(40):

$$D(gbest(t-1), P(t)) = \begin{cases} gbest(t), & \text{if } f(gbest(t)) < f(gbest(t-1)) \\ gbest(t-1) & \text{otherwise} \end{cases} \quad (40)$$

in which we also have:

$$f(gbest(t)) = \min(f(pbest_1(t), f(pbest_2(t), \ldots, f(pbest_N(t))))$$

in the $t^{th}$ generation.

We also consider the 'best' case first in which for each particle $i$, the $pbest$ of every particle could always be updated. Then we have:

$$pbest_{i,d}(t) = x_{i,d}(t) \quad (41)$$

Luckily the proof of the dynamic system constructed by Eq. (34)-(35) converges to equilibrium was given in [56], claiming that $x_{i,d}(t)$ can guarantee convergence (In our discussed case,

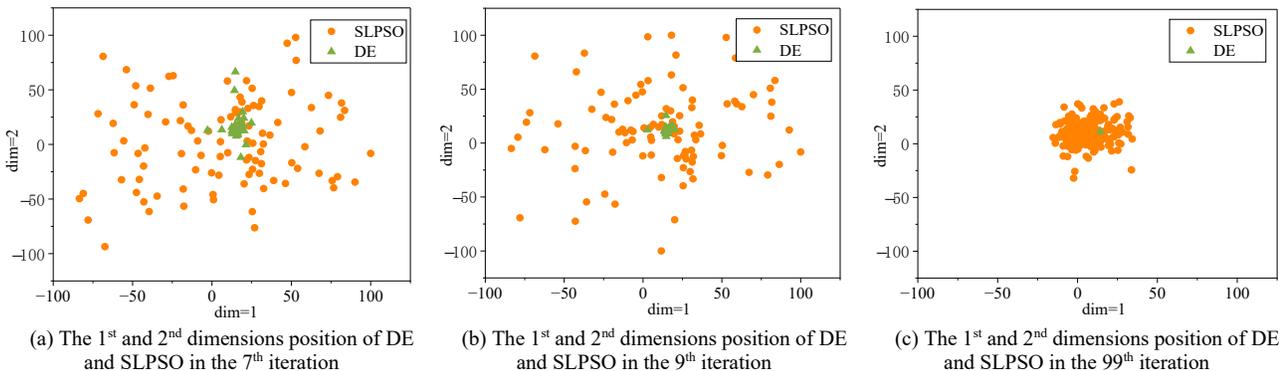

(a) The 1st and 2nd dimensions position of DE and SLPSO in the 7th iteration
(b) The 1st and 2nd dimensions position of DE and SLPSO in the 9th iteration
(c) The 1st and 2nd dimensions position of DE and SLPSO in the 99th iteration

Fig. 7. The 1st and 2nd dimensions position of DE and SLPSO on the early stage. (a) in the 7th iteration; (b) in the 9th iteration; (c) in the 99th iteration.

it means that $pbest_{i,d}(t)$ guarantees to convergence), we therefore do not need to further prove it is stable convergence.

According to **Theorem 3**, we can easily conclude that SLPSO cannot guarantee global convergence.

With the SDMC method, since it is proved that when $t \to \infty$, $v(C_i(t)) \to 0$ for $i \in [1, N]$, leading to a consequence that $\lim_{t \to \infty} v(M(t)) = 0$ and thus there must have a time *t'* after which $v(U_{t',+\infty}) < v(S)$ since $v(M(t))$ tends to 0. Then **(H5)** could not hold. Therefore, SLPSO could not guaranteed to converge to the global optimum.

Although SLPSO could not guaranteed to converge to the global optimum, it still has better diversity than DE. Let's start with Eq.(34), which can be written as:

$$v_{i,d}(t+1) = r_1 v_{i,d}(t) + r_2(x_{gbest,d}(t) - x_{i,d}(t)) \\ + r_2(x_{k,d}(t) - x_{k2,d}(t)) + \varepsilon r_3(\overline{x_d}(t) - x_{i,d}(t)) \quad (42) \\ + r_2(x_{k2,d}(t) - x_{gbest,d}(t))$$

where *k2* is a randomly selected particle in the population.

We rewrite the updated formula in this way so we can split the directionality and randomness parts of $x_{k,d}(t) - x_{i,d}(t)$, since the distribution of the direction $x_{k2,d}(t) - x_{gbest,d}(t)$ is no longer related to the adaptation value of the particle *k2*, but to the value of $x_{gbest,d}(t)$.

In this way, we can compare SLPSO with DE and find that in Eq. (42), both the historical velocity $r_1 v_{i,d}(t)$, the part that learns from the mean position in the population $\varepsilon r_3(\overline{x_d}(t) - x_{i,d}(t))$ and $r_2(x_{k2,d}(t) - x_{gbest,d}(t))$ that leads *gbest* to learn from a random particle can both slow down the speed at which the *gbest* particles stall at one point. Obviously, this will also benefit the diversity of the entire swarm. In addition, for other particles, the addition of the hyperparameter $\varepsilon$ making $\varepsilon r_3(\overline{x_d}(t) - x_{i,d}(t))$ (the only part that can help particles to aggregate) cannot counteract the effect of other terms on particle diversity, which will also help SLPSO aggregate more slowly compared with DE.

Our conclusion is validated by Fig. 7, which shows the performance comparison between SLPSO and DE on the sphere function with 1000 dimensions in the early stage. We find that DE converges to a very small interval very early on (in the 99[th] generation), while SLPSO is still searching in a relatively large range.

*4) Analysis of GTPSO*

For GTPSO, both GT in Eq. (38) and Eq. (39) can help the *gbest* particle to have a new $C_{gbest}(t)$ with positive measure even in the situation that SLPSO can only have $v(C_{gbest}(t)) = 0$.

In addition, we believe that in GTPSO, the GT with Eq. (39) method can bring better improvements to the algorithm. Imagine the following scenario: the *gbest* particle has already stagnated ($v_{gbest,d}(t) = 0$) in the *t* generation, and it happens that all particles have gathered at that point from this generation on, resulting that there must have $v(M(t+1)) = 0$ with the use of Eq.(38). However, at this point, since all particles have only begun to stagnate from this generation, we must have $\exists v_{i,d}(t) \neq 0, i \in [1, N]$, resulting $v_{gbest,d}(t+1) \neq 0$ and thus $v(M(t)) > 0$. In fact, this is only a hypothesis for the most extreme case, and more often Eq. (39) can delay this scenario from happening.

The ablation experiments given in [45] compares the different GT strategies in GTPSO. These results can also support the above statement since in most of the functions, GT with Eq. (39) brings a greater boost to SLPSO compared with GT with Eq. (38).

However, GTPSO is not an algorithm that guarantees to converge to the global optimum as **(H5)** is not guaranteed to hold even in the 'best' case. In GTPSO, the standard deviation of the Gaussian distribution tends to 0 when $v_{i,d}(t) \to 0$ for the entire swarm, indicating $v(M(t)) \to 0$.

The GTPSO still outperforms GTDE in most of the benchmark functions, showing that whether the global convergence in **THEOREM 1** is satisfied is not the only correlation factor with the final performance of the algorithm. This is because the maximum evaluation number is not infinite under the practical circumstance. although improvements based on this can improve the performance of algorithms in many cases.

We show the standard deviation of the 1[st] dimension of the particles in SLPSO and GTPSO on sphere function with DIM=1000 at the early stage and the later stage in Fig. 8 (a) and (b), respectively. In Fig. 8, we can observe that GT greatly slows down the aggregation of all the particles in the GTPSO to a point, comparing to SLPSO, providing opportunities to jump out of this stagnant state in the medium term. Hence, there is more time for the GTPSO to search effectively.

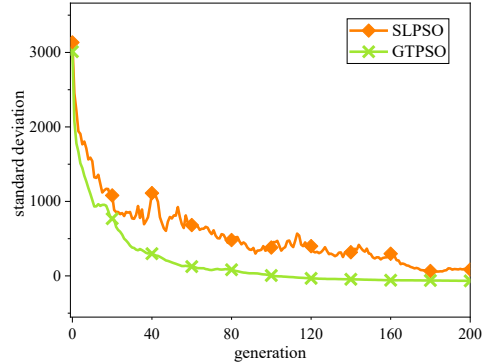

(a) On the early stage

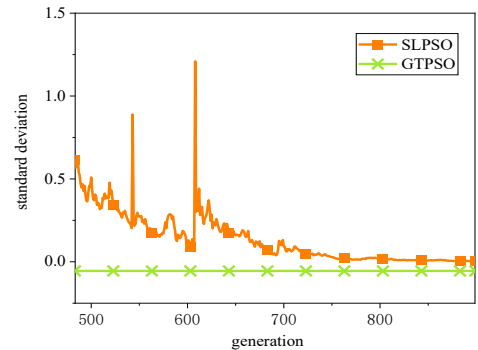

(b) On the later stage

Fig. 8. The standard deviation of the 1[st] dimension of the individuals in SLPSO and GTPSO on sphere function with DIM=1000 in different stages (a) On the early stage; (b) On the later stage.

Afterwards, as we mentioned before, since the GTPSO does not satisfy **(H5)**, it also inevitably falls into the aggregation state when *t* becomes large.

### C. Suggestions of Integrating GT in Algorithm Design

Based on the results of our analysis and the above observations, we recommend the following suggestions when designing and integrating GT:

1. Observing how the algorithm behaves on the sphere function, as the function has some good properties and provides the same 'importance' in different dimensions.
2. Analyzing the global convergence of the algorithm, and comparing with the properties of the characteristics of basic algorithms of the proposed GT variant. We suggest applying GT on algorithms that converge relatively fast, but such operation will also result in inevitable additional fitness evaluation.
3. After the above analysis, while designing GT, we suggest to rationally use distributions that can always cover the entire search scope during the whole evolution, e.g., Gaussian distributions with $\sigma > 0$ all the time or uniform distribution covering the search scope to enhance the global convergence (i.e., diversity) of the algorithm. For the algorithms that are already globally convergent, it is better to design GT that accelerates the concentration of individuals, such as learning from a particular individual (usually the best individual or the mean of individual positions).
4. For the choice of 'target gene', we recommend integrating GT with individuals that have a greater influence on others as well as a relatively small measure for its search space (that is, a small $C_i(t)$), e.g., the *gbest* in SLPSO and the best individual in DE, which can better guide the search with relatively fewer drawbacks.
5. For different situations, the properties required by the algorithm at different stages may be different. Observing the integrated algorithm and using GT at a specific stage may improve the performance of the algorithm, which has not been investigated by GTDE or GTPSO.

### D. Experimental Studies

Based on all those suggestions, we make an improvement on GTPSO to validate the practicability of our statement. That is, changing the standard deviation of the Gaussian distribution in Eq.(39) from $\frac{1}{2}(v_{k_1,d}(t) - v_{k_2,d}(t))$ to a constant value $\sigma$, allowing the GTPSO to satisfy **(H5)**. Then Eq. (39) becomes:

$$v_{i,d}(t+1) = Gaussian(\frac{1}{2}(v_{k_1,d}(t) + v_{k_2,d}(t)), \sigma) \quad (43)$$

Table I shows the performance of the original GTPSO and its variants with different $\sigma$ values. We also analyze the sensitiveness of $\sigma$ by setting it to different values, i.e., $\sigma = 0.1$, 1, and 10. The experiments run on the same benchmark functions as in the original GTPSO, and we do not make any changes to any other hyperparameters. Each test is executed 30 times independently. The '+', '≈' and '–' represent that the improved GTPSO variant is significantly better, worse, and similar to the original GTPSO.

It can be seen that this small change greatly improves the effect of GTPSO in the vast majority of functions, whether in the case of $\sigma = 0.1$, 1, or 10. However, as the value of $\sigma$ gets larger, the less the increase provided by the new constant $\sigma$ to GTPSO becomes. Therefore, we recommend adopting GT in the situation that when the measure of the possible searching space of the swarm/population becomes smaller, a search more focused on a small area would be more conducive for GT.

From the results, we therefore believe that these 12 benchmark functions require higher global convergence for GTPSO, and the experimental results validate the recommendations we propose in this section.

TABLE I
EXPERIMENTAL RESULTS FOR THE ORIGINAL GTPSO AND THE IMPROVED GTPSO VARIANTS ON THE 12 BENCHMARK FUNCTIONS

| Function | GTPSO | GTPSO-$\sigma$=0.1 | GTPSO-$\sigma$=1 | GTPSO-$\sigma$=10 |
|---|---|---|---|---|
| | Mean±Std | Mean±Std | Mean±Std | Mean±Std |
| $F_1$ | 8.22E+01±3.88E+02 | **4.08E–04±7.33E–05** (+) | 1.43E–02±4.46E–03 (+) | 6.81E–01±2.38E–01 (+) |
| $F_2$ | 4.57E+05±1.07E+05 | **2.01E+05±1.44E+04** (+) | 4.95E+05±2.07E+04 (–) | 1.64E+06±3.86E+04 (–) |
| $F_3$ | 8.75E+01±4.76E+00 | 3.68E+01±1.82E+00 (+) | **2.81E+01±1.83E+00** (+) | 2.92E+01±1.98E+00 (+) |
| $F_4$ | 1.77E+02±1.28E+02 | 4.03E+00±2.07E+00 (+) | **0.00E+00±0.00E+00** (+) | 3.20E+00±1.33E+00 (+) |
| $F_5$ | 1.67E+00±5.91E–01 | **7.88E–01±4.96E–02** (+) | 3.64E+00±3.86E–01 (–) | 4.51E+00±1.79E-01 (–) |
| $F_6$ | 3.40E+03±1.40E+03 | **2.80E+00±5.55E–01** (+) | 4.71E+02±4.03E+01 (+) | 7.72E+03±4.87E+02 (–) |
| $F_7$ | 1.33E+05±4.18E+04 | 1.03E+05±2.92E+03 (+) | 1.03E+05±3.90E+03 (+) | **1.01E+05±1.66E+03** (+) |
| $F_8$ | **7.49E+02±1.57E+02** | 1.61E+03±7.50E+01 (–) | 1.58E+02±7.80E+00 (+) | 5.49E+02±1.08E+01 (–) |
| $F_9$ | 6.32E–01±6.53E–01 | **1.84E–03±1.87E–04** (+) | 9.94E–03±1.30E–03 (+) | 2.33E–01±1.87E–02 (+) |
| $F_{10}$ | 1.97E–01±2.81E–01 | 5.77E–04±2.18E–03 (+) | **4.98E–05±8.18E–05** (+) | 1.67E–03±4.37E–04 (+) |
| $F_{11}$ | 4.39E+00±1.61E+00 | **3.11E–04±9.33E–04** (+) | 3.57E–04±9.89E–04 (+) | 1.33E–03±1.22E–03 (+) |
| $F_{12}$ | 1.02E+09±1.43E+08 | 9.27E+08±1.50E+08 (+) | **4.25E+08±6.96E+07** (+) | 1.29E+09±9.15E+07 (≈) |
| Number of +/≈/– | | 11/0/1 | 10/0/2 | 7/1/4 |

For finer designs, we recommend associating this parameter $\sigma$ with the search scope of the function, rather than giving the same constant to functions with all different scopes.

## VI. CONCLUSION

In this article, we briefly introduce and compare two types of convergence (i.e., stable convergence and global convergence) that are distinct and even fundamentally opposite in meaning. We then prove that individual convergence and population convergence in stable convergence are mutually exclusive with global convergence. Inspired by the mutual exclusion, we propose the simple yet complete SDMC method to determine whether an algorithm can converge to the global optimum by comparing the measure of the search scope of individuals at each moment with the measure of the feasible domain of the problem. We also provide a proof that our newly proposed **(H5)** is a necessary and sufficient condition for global convergence.

To demonstrate that the proposed SDMC method is more accurate and comprehensive than the commonly used method in the EC field, we analyze LDIW-PSO and a simple periodic partitioned uniform sampling method as examples to show that SDMC is more general for global convergence analysis.

Subsequently, to explore the assistance of theoretical research in algorithm design, we analyze the two GT-based algorithms, GTDE, and GTPSO, and their original algorithms, DE and SLPSO, by using the SDMC method. We discuss the role of GT in the global convergence of large-scale optimization algorithms. Following this, we propose some suggestions on how to assist in designing algorithms that guarantee global convergence and use a simple improvement to demonstrate the practicality of these suggestions.

Moreover, during the analysis and comparison of the above two groups of algorithms, we find that comparing the measure of search scope can somehow compare the strength of global convergence properties across different algorithms, the comparison of two groups of algorithms that do not guarantee global optimality confirms our conjecture. In the future, we plan to provide corresponding theoretical foundations to support the above assumption and to compare advanced algorithms as well as analyze the impact of the global convergence property on their performance across different problems.